\documentclass{article}

\PassOptionsToPackage{numbers, compress}{natbib}
\usepackage{flafter}

\usepackage[final]{neurips_2021}

\usepackage{graphicx}

\usepackage[utf8]{inputenc} 
\usepackage[T1]{fontenc}    
\usepackage{hyperref}       
\usepackage{url}            
\usepackage{booktabs}       
\usepackage{amsfonts}       
\usepackage{nicefrac}       
\usepackage{microtype}      
\usepackage{xcolor}         
\usepackage{multirow}

\title{Interpreting Language Models Through Knowledge Graph Extraction}

\author{Vinitra Swamy \qquad Angelika Romanou  \qquad Martin Jaggi\\
  École Polytechnique Fédérale de Lausanne (EPFL) \\
  Lausanne, Switzerland \\
  \texttt{firstname.lastname@epfl.ch} \\}

\begin{document}

\maketitle

\begin{abstract}
Transformer-based language models trained on large text corpora have enjoyed immense popularity in the natural language processing community and are commonly used as a starting point for downstream tasks. While these models are undeniably useful, it is a challenge to quantify their performance beyond traditional accuracy metrics. In this paper, we compare BERT-based language models through snapshots of acquired knowledge at sequential stages of the training process. Structured relationships from training corpora may be uncovered through querying a masked language model with probing tasks. We present a methodology to unveil a knowledge acquisition timeline by generating knowledge graph extracts from cloze "fill-in-the-blank" statements at various stages of RoBERTa's early training. We extend this analysis to a comparison of pretrained variations of BERT models (DistilBERT, BERT-base, RoBERTa). This work proposes a quantitative framework to compare language models through knowledge graph extraction (GED, Graph2Vec) and showcases a part-of-speech analysis (POSOR) to identify the linguistic strengths of each model variant. Using these metrics, machine learning practitioners can compare models, diagnose their models' behavioral strengths and weaknesses, and identify new targeted datasets to improve model performance.
\end{abstract}

\section{Introduction}
The evolution of deep learning methodologies and continual expansion of computing capacity has enabled many advancements in language modeling field. In specific, transformer-based architectures have foreshadowed the creation of BERT and its many variants, surpassing previously held records in GLUE, SQuAD, and MultiNLI benchmarks \cite{devlin2018BERT, wang2018glue, rajpurkar2016squad, williams2017broad}. BERT-based architectures have become lightweight and more efficient (DistilBERT) and trained more effectively to become increasingly performant (RoBERTa) \cite{sanh2019distilBERT, liu2019RoBERTa}.

As we build more capable machine learning models with increasingly minute differences in performance benchmarks, it is important that our model behavior is interpretable. Evaluation of language model performance is moving towards fine-grained diagnostics that go beyond simply answering if a model outperforms another and instead delve into the scale and type of errors the model makes. There are two main approaches in this area: curating a set of difficult examples, and developing in-depth metrics for targeted model evaluation \cite{MLandNLP37:online}. Our work is positioned within the latter approach, as a diagnostic benchmark to allow for nuanced studies into a model's strengths and weaknesses. These diagnostic studies, often using counterfactual or annotated training examples, can be followed by tailored fine-tuning approaches to improve deficient areas \cite{fu-etal-2020-interpretable, Kaushik2020Learning, gardner2020evaluating, warstadts_doi}. 

In this paper, we contribute a pipeline for deeper analysis of language model performance by generating knowledge graph (KG) extracts as inspired from  \citet{petroni2019language}. We aim to address two main research directions: 
\begin{itemize}
    \item How can we quantitatively compare language model knowledge acquisition? Can we extend this analysis to the same model at different stages of early training?
    \item As a language model trains, what linguistic traits does it learn over time?
\end{itemize}
We run this pipeline on a RoBERTa architecture at various stages of early training and on pretrained models from the Transformers library (BERT-Base, DistilBERT, RoBERTa) \cite{wolf-etal-2020-transformers}. Our objective is to capture a snapshot of learned knowledge from the current state of the model. We achieve this by training with a masked language model objective and querying our models using cloze "fill-in-the-blank" statements. After replacing the predicted word in the masked statement, we generate a knowledge graph comprised of subject-verb-object triples which acts as a representation of knowledge generated from the language model. We analyze this work in two ways: by extending previous literature in graph representations to quantitatively compare differences between language model knowledge extracts using graph-edit distance and graph2vec metrics and by proposing a part-of-speech (POS) tagging analysis to better understand a model's linguistic strengths. The novelty in this work arises from using these proposed metrics to compare language models with each other over time, in an analysis that goes beyond accuracy and loss. The results of our approach show strong evidence for how language models learn knowledge through different training epochs and variants, increasing the interpretability of language model performance. We release the code for our experiments in our Github repository \footnote{https://github.com/epfml/interpret-lm-knowledge}.

\section{Related work}

The rise of deep learning architectures in language modeling has been followed closely by explainability research seeking to understand what these models encode. BERT is successful at encoding information at the syntactic and semantic level \cite{rogers2020primer}. \citet{hewitt-manning-2019-structural} proposes a probe that identifies whether syntax trees can be represented as a linear transformation of BERT embeddings. \citet{clark2019does} uses probing classifiers to demonstrate that a number of BERT attention heads correspond to syntactic tasks, while \citet{goldberg2019assessing} demonstrates BERT's capabilities to solve syntax-based objectives, like subject-verb agreement. \citet{coenen2019visualizing} focuses on BERT's attention matrices, examining dependency relations represented as certain directions in the matrix subspace. \citet{coenen2019visualizing} further suggests that BERT’s internal geometry can be represented as separate linear subspaces for syntactic and semantic information. It is evident that BERT encodes knowledge beyond language representations into the syntactic and semantic space, but the full extent of this is still debated upon.

Several recent studies divide the model into smaller architecture blocks such as layers, encoded hidden states, and attention heads to expose where in the model specific information is being stored. \citet{tenney2019BERT} shows that the top layers of BERT address long-range dependencies (subject-object dependencies) while the early layers of BERT encode short dependency relationships at the syntactic level (subject-verb agreement). \citet{peters-etal-2018-deep} extends this conclusion to Convolutional, LSTM, and self-attention architectures. \citet{hao2019visualizing} implies that layers close to inputs learn more transferable language representations by exposing which layers of BERT change during finetuning. Our novelty lies in identifying when certain knowledge is acquired in the model training process instead of aiming to find where in the model the information is stored. Recent papers from \citet{liu2021probing, chiang2020pretrained} analyze when a language model gains certain types of knowledge (commonsense, factual), but focus on case studies of individual language models instead of comparisons across language models over time.

In recent work towards understanding language model errors, training examples are annotated and used to probe for model behavioral phenomena common to a task of interest. Minimal pairs, also known as counterfactual examples or contrast sets, perturb examples and often change the gold label to highlight model weaknesses \cite{Kaushik2020Learning, gardner2020evaluating, warstadts_doi}. \citet{ribeiro-etal-2020-beyond} proposes a CheckList framework, which enables the semi-automatic creation of such test cases. Alternatively, \citet{fu-etal-2020-interpretable} shows that examples can be annotated with different attributes, which allow for a more granular analysis of missed areas. These approaches seek to augment existing data for insights into model performance.

The most relevant area to our approach is in knowledge graph extraction from language models. LAMA, released by Petroni et. al, is a pipeline from Facebook to extract knowledge bases directly from language models \cite{petroni2019language, petroni2020how}. The LAMA methodology focuses on extracting relation triples from cloze statements to generate knowledge graphs from language models. While we are heavily inspired by this pipeline and re-use the cloze statement datasets presented in LAMA, their work fails to consider the extracted knowledge graphs in an interpretability context, focusing on the quality of KG extraction instead of viewing the differences between the graphs as a diagnostic metric for language model performance. We also use a simpler and faster relation triple extraction methodology through natural language processing libraries instead of LAMA's attention-based approach, as we are more interested in the evaluation of these graphs than the accuracy of their creation. Recent papers by \citet{heinzerling2021language} and \citet{jiang2020can} confirm the claims of LAMA and review two more methodologies to extract knowledge graphs from language models. \citet{aspillaga2021tracking} uses query-answering probing tasks and compare knowledge graph concept relatedness to WordNet. \citet{goswami2020unsupervised} propose ReFLEX, which focuses on unsupervised cloze template completion using language models, but does not extend the work towards knowledge graph analysis.

In summary, the relevant literature shows that BERT and its transformer-based variants do store information at the syntactic and semantic level. This demonstrates that there is some signal to capture in our probing experiments. In related explainability literature, the work focuses on revealing which blocks of a transformer-based model are responsible for storing a specific type of information with moderate success. We extend this analysis to understand when certain knowledge is acquired over time in the model training process, comparing across multiple language model variations. Diagnostic benchmarks for language models, aiming to demonstrate model strengths and weaknesses are more in line with the purpose of this work, but focus mainly on augmenting data examples instead of measuring knowledge. The literature on knowledge graph extraction is closest to our work, specifically \citet{petroni2019language}'s approach to treating language models as knowledge bases. We propose a novel extension of LAMA by using a simpler relation extraction methodology and extending metrics from graph literature \cite{grapheditdistance, narayanan2017graph2vec} to analyze knowledge graph extracts over time across multiple models. Probing task literature helps us contextualize evaluation strategies, namely the benchmarks presented in \citet{kim-etal-2019-probing}'s work to identify skills in a language model, like predicting the correct wh- word (why, when, where) and identifying the correct coordinating conjunction. We explore aspects of this linguistic analysis in our results section with grammatical tagging, but future work involves a more granular exploration of these probing task benchmarks on our pipeline.

\section{Methodology}

\begin{figure*}[t]
    \centering
    \includegraphics[width=1\textwidth]{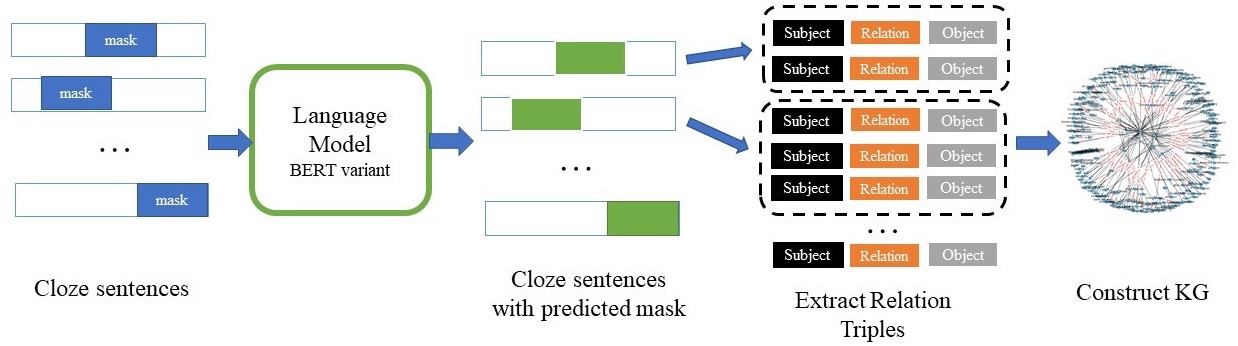}
    \caption{\label{fig:normal_case}Architecture of our probing pipeline to generate knowledge graphs from language models.}
\end{figure*}

To extract knowledge graphs from a language model, we begin by obtaining the model in question (either training early epochs from scratch as referenced in implementation details below, or loading the pretrained model from an existing library) using a standard masked language model objective. We query the model using cloze statements from our 4 datasets, i.e. "During Super Bowl 50 the [MASK] gaming company debuted their ad for the first time." from the LAMA SQuAD dataset where the intended gold label is "Nintendo" \cite{petroni2019language}. The language model predicts the top 5 missing masks, we select the one that is most likely (based on output probability) and replace it in the sentence. We then extract relevant subject-relation-object triples from these sentences and construct a knowledge graph. Figure \ref{fig:normal_case} represents our knowledge graph extraction pipeline. When we reference Ground Truth Knowledge graphs, this is simply a baseline approach using the SpaCy and Textacy libraries to extract triples from the gold-label sentences \cite{spacy, textacy47:online}.

While deciding upon an optimal method for relation extraction, we explore two different natural language processing libraries: SpaCy and Textacy. The SpaCy approach uses a dependency parser to identify entities and extract the relevant subject-relation-object triples in our sentences, while Textacy abstracts the process into their own library. We noticed that these out-of-the-box taggers were failing to capture certain triples that made intuitive sense, so we extend the SpaCy implementation by creating a hybrid approach with linguistic rule-based modifications. Our hybrid implementation provides many more triples, including a number that out-of-the-box methodologies from Textacy and SpaCy do not recognize, as highlighted in Table \ref{tab:graph_results}. An example of subject-relation-object triple extraction for the sentence "The ad shown during the Super Bowl for the next Jason Bourne movie was paid by Sony." is the following list of tuples: \{\textit{ad, shown by, The Super Bowl}\}, \{\textit{ad, paid by, Sony}\}.

\begin{table}
\centering

\label{tab:graph_results}
\begin{tabular}{lcc} 
\hline
\multicolumn{1}{c}{\textbf{Sentence}}                                                                                                                                                    & \textbf{Extractor} & \textbf{Triple}                                                                                             \\ 
\hline
\multirow{3}{*}{\begin{tabular}[c]{@{}l@{}}The ad shown during the Super Bowl \\ for the next Jason Bourne movie was \\ paid by [Sony].\end{tabular}}                                    & SpaCy              & {[}]                                                                                                        \\ 
\cline{2-3}
                                                                                                                                                                                         & Textacy            & {[}{[}`ad', `was paid', `Sony']]                                                                            \\ 
\cline{2-3}
                                                                                                                                                                                         & Custom             & \begin{tabular}[c]{@{}c@{}}[[`ad', `shown by', `The Super Bowl'],\\{[}`ad', `paid by', `Sony']]\end{tabular}  \\ 
\hline
\multirow{3}{*}{\begin{tabular}[c]{@{}l@{}}In Groner v Minister for Education, \\the Court of Justice accepted \\ {[}Gaelic] to be required to teach \\in Dublin colleges.\end{tabular}} & SpaCy              & \begin{tabular}[c]{@{}c@{}}{[}{[}`the Court of Justice', `accepted in', \\`Groner']]\end{tabular}           \\ 
\cline{2-3}
                                                                                                                                                                                         & Textacy            & \begin{tabular}[c]{@{}c@{}}[[`Gaelic', `to be required', \\`to teach in Dublin colleges']]\end{tabular}       \\ 
\cline{2-3}
                                                                                                                                                                                         & Custom             & {[}{[}`Dublin colleges', `teach', `Gaelic']]                                                                \\
\hline
\end{tabular}
\vspace{2mm}
\caption{\label{tab:graph_results}Subject-relation-object extraction examples between out-of-the-box SpaCy and Textacy taggers and our custom parser.}
\end{table}

\subsection{Model Training details}
 We use PyTorch for our training model code, aided by the HuggingFace Transformers library \cite{wolf-etal-2020-transformers}. In preprocessing our evaluation data, we use a LineByLineTextTokenizer from Transformers. 
 
 As mentioned in the previous sections, the models we use for evaluation are standard BERT-inspired architectures. More precisely, we have three variations of pretrained architectures: BERT, DistilBERT, and RoBERTa \cite{devlin2018BERT, sanh2019distilBERT, liu2019RoBERTa}. The reason for choosing these specific models is because of their immense popularity and their ability to show an evolution in training, where DistilBERT is a smaller approximation of BERT trained for less time, BERT-base is very widely used and is our standard reference point, and RoBERTa is built on top of BERT, trained for a longer amount of time and with more data. We utilize a pretrained BERT-based-cased, with 12 layers, 12 attention heads and 109M parameters, trained on cased English text from Wikipedia. For RoBERTa, we utilize a pretrained model with a similar architecture modifying BERT-base, with 12 layers, 12 attention heads, and 125M parameters, trained on 16GB of uncompressed English text from BookCorpus and English Wikipedia \cite{liu2019RoBERTa}. DistilBERT was also pretrained and originally distilled from the BERT-base-cased checkpoint, using 6 layers, 768 hidden nodes, 12 attention heads and 66M parameters. All results can be replicated using pretrained models from version 4.2.0 of Transformers \cite{wolf-etal-2020-transformers} and a standard NVidia K80 / T4 GPU.

For the training experiments, we train RoBERTa from scratch, for increments of 1, 3, 5, and 7 epochs on a Tesla P100 GPU with 52,000 vocab size, 514 position embeddings, 12 attention heads, batch size of 64, and 6 hidden layers from 15 GB of English Wikipedia articles \footnote{We use an English Wikipedia extract file, enwiki-latest-pages-articles, from November 2020. A script with further replication details can be found in our Github repository, using the smaller sample corpus (enwiki-10) with 0.5 GB of data.}. We use the Transformers library's ByteLevelBPETokenizer to preprocess the data with block size 128, masking probability 0.15, and train RoBERTa with the standard masked language model training objective \cite{liu2019RoBERTa}. 

\section{Experiments}

We begin by highlighting our datasets, introduced by Petroni et al. in LAMA \cite{petroni2019language, petroni2020how}. We then describe our evaluation metrics and probing tasks, and present our experimental KG results.

We use two cloze "fill-in-the-blank" statement datasets from LAMA to query our language models and generate knowledge graphs \cite{petroni2019language}. These datasets are based on the Stanford Question Answering Dataset (SQuAD) and the Google Relation Extraction Corpus (Google-RE) \cite{rajpurkar2016squad, GoogleAI29:online}. Both datasets are derived from Wikipedia, the same data that was used to train BERT and its variants, which is why we can use them to query information the language model has been exposed to.

\subsubsection{SQuAD}
SQuAD \cite{rajpurkar2016squad} is widely used in the field of question-answering. Petroni et al. select a subset of 305 questions that are context-insensitive from the SQuAD development set, and manually convert the question-answer pairs to cloze-style questions \cite{petroni2019language}. For example, the question “Who developed the theory of relativity?” is rewritten as “The theory of relativity was developed by [MASK]”. The SQuAD cloze statement dataset is the smallest of the four LAMA datasets.

\subsubsection{Google-RE}
The Google-RE corpus contains over 60,000 facts extracted from Wikipedia. Petroni et al. use three of Google-RE's five databases, focusing on facts pertaining to place of birth, date of birth, and place of death \cite{petroni2019language}. Each fact is directly tied to a Wikipedia text supporting it. Each sentence is sampled with a template sentence format; for the place of birth dataset, sentences that fit the template "[S] was born in [O] ..." were extracted. The three Google-RE cloze datasets (place-of-birth, date-of-birth, place-of-death) are significantly larger than the SQuAD cloze dataset, with 2937, 1825, and 768 statements respectively.

\begin{table*}[tbp]
    \centering
    \small
    \begin{tabular}{lcc}
    \hline
    \textbf{Target Model} & \textbf{\begin{tabular}[c]{@{}c@{}}Graph edit distance\\ on the extracted knowledge graph\end{tabular}} & \textbf{\begin{tabular}[c]{@{}c@{}}Euclidean distance \\ on the graph2vec embeddings\end{tabular}} \\ 
    \hline
    \textbf{RoBERTa 1e}     & 141.25 & 0.2260 \\ 
    \textbf{RoBERTa 3e}    & 135.00 & 0.1733 \\ 
    \textbf{RoBERTa 5e}    & 130.50 & 0.1607 \\ 
    \textbf{RoBERTa 7e}    & 121.50 & 0.1605 \\ 
    \textbf{DistilBERT}          & 28.50 & 0.0284 \\ 
    \textbf{BERT}                & 16.50 & 0.0202 \\ \hline
    \end{tabular}
    \caption{\label{tab:graph_results}Distance from the target models to pretrained RoBERTa using the graph-edit-distance and graph2vec metrics on the Google RE Place of Birth dataset.}
\end{table*}

\subsection{Evaluation Metrics}
To obtain a quantitative comparison of our knowledge graph extracts, we present two approaches from related literature, graph-edit-distance and graph2vec \cite{grapheditdistance, narayanan2017graph2vec}.  These metrics can be used to compare the extracted knowledge graph with the ground truth knowledge graph to measure accuracy. However, more novelty arises in using these metrics to compare language models against each other, as shown later in the results where BERT, DistilBERT, and the trained from scratch variants are compared with pretrained RoBERTa.

The first approach, graph-edit-distance (GED), is also known as tree-edit-distance for rooted trees \cite{grapheditdistance}. GED is a measure for the similarity between two graphs and is commonly used in knowledge graph literature. As we are not measuring hierarchical relationships, we choose this metric to compare our graphs. Given a standard set of graph edit operations (vertex insertion, vertex deletion, vertex substitution, edge insertion, edge deletion, edge substitution), the graph edit distance between two graphs $g_{1}$ and $g_{2}$ is defined as
$$ GED(g_{1}, g_{2}) = \min_{(e_{1},...,e_{k}) \in \mathcal{P}(g_{1},g_{2})} \sum_{i=1}^{k} c(e_{i})$$
where $\mathcal{P}(g_{1},g_{2})$ denotes the set of edit paths transforming $g_{1}$ into (a graph isomorphic to) $g_{2}$ and $c(e)\geq 0$ is the cost of each graph edit operation $e$. In our implementation of GED, $c(e)=1$ for every graph edit operation. Since our knowledge graphs are large and our compute is limited, we use an approximation of graph edit distance (Assignment Edit Distance) as presented by \citet{riesen2009approximate}.

In our second quantitative evaluation metric, we transform the knowledge graphs into embeddings and compare these graph vector representations using Euclidean distance. 
We utilize the graph2vec functionality from the KarateClub \footnote{https://karateclub.readthedocs.io/} library and train a FEATHER graph level embedding \cite{rozemberczki2020characteristic} with the standard hyperparameter settings to embed each language model's knowledge graph into a 100-dimensional vector space. In order to increase the connectivity of the graph, we stem the tokens in each of the subject-verb-object triples. A future extension could be to apply lemmatization instead, using the part-of-speech tag of the token.

In addition to the graph evaluation metrics, we investigate how representations of linguistic structure are learned over time. To understand the grammatical construction of sentence, interpret its underlying meaning, and extract subject-verb-object relationships, part-of-speech tagging is an important step as showcased by \citet{saphra2018language, manning2011part, voutilainen2003part} among many others. We extend related literature in probing tasks \cite{kim-etal-2019-probing} by proposing a new diagnostic linguistic metric using part-of-speech tagging (POS-tags) on the relation triples with Natural Language Toolkit (NLTK)'s Average Perceptron Tagger for the English language \cite{nltk}. This POS-tag percentage difference metric compares the number of times each part of speech appears in the relations of the language model's knowledge graph versus the ground truth knowledge graph. The motivation for this analysis is in allowing us to understand how a language model learns in relation to different parts of speech. We define this metric as the POSOR (Part-of-Speech Overprediction Rate) and represent it with the following equation
$$ POSOR(pos) = \frac{(LM_{pos} - GLM_{pos})\cdot 100}{GLM_{pos}} $$
where $LM_{pos}$ is the number of triples extracted from the language model knowledge graph for each POS category, and $GLM_{pos}$ is the number of triples extracted from the ground truth knowledge graph for each POS category.

We first tag each relation triple with the appropriate POS-tags across both knowledge graphs, compute the difference in the counts of POS-tags appearing in each graph, and then normalize over the ground truth counts to create POS-tag percentage differences (Table \ref{ref:squad_pos_results}). This allows us to suggest statements of the format: Model A overpredicts nouns (high positive percentage difference) when it should be predicting more adverbs (high negative percentage difference). Analyzing models through these statements is useful because it enables a data scientist to select a relevant subset of training data to fix deficiencies in a language model's performance. The NLTK POS tagset consists of verbs, nouns, pronouns, adjectives, adverbs, adpositions (prepositions and postpositions), conjunctions, determiners, cardinal numbers, particles or other function words, and punctuation \cite{nltk}. 

\begin{figure*}[t]
  \centering
  \includegraphics[width=0.93\linewidth]{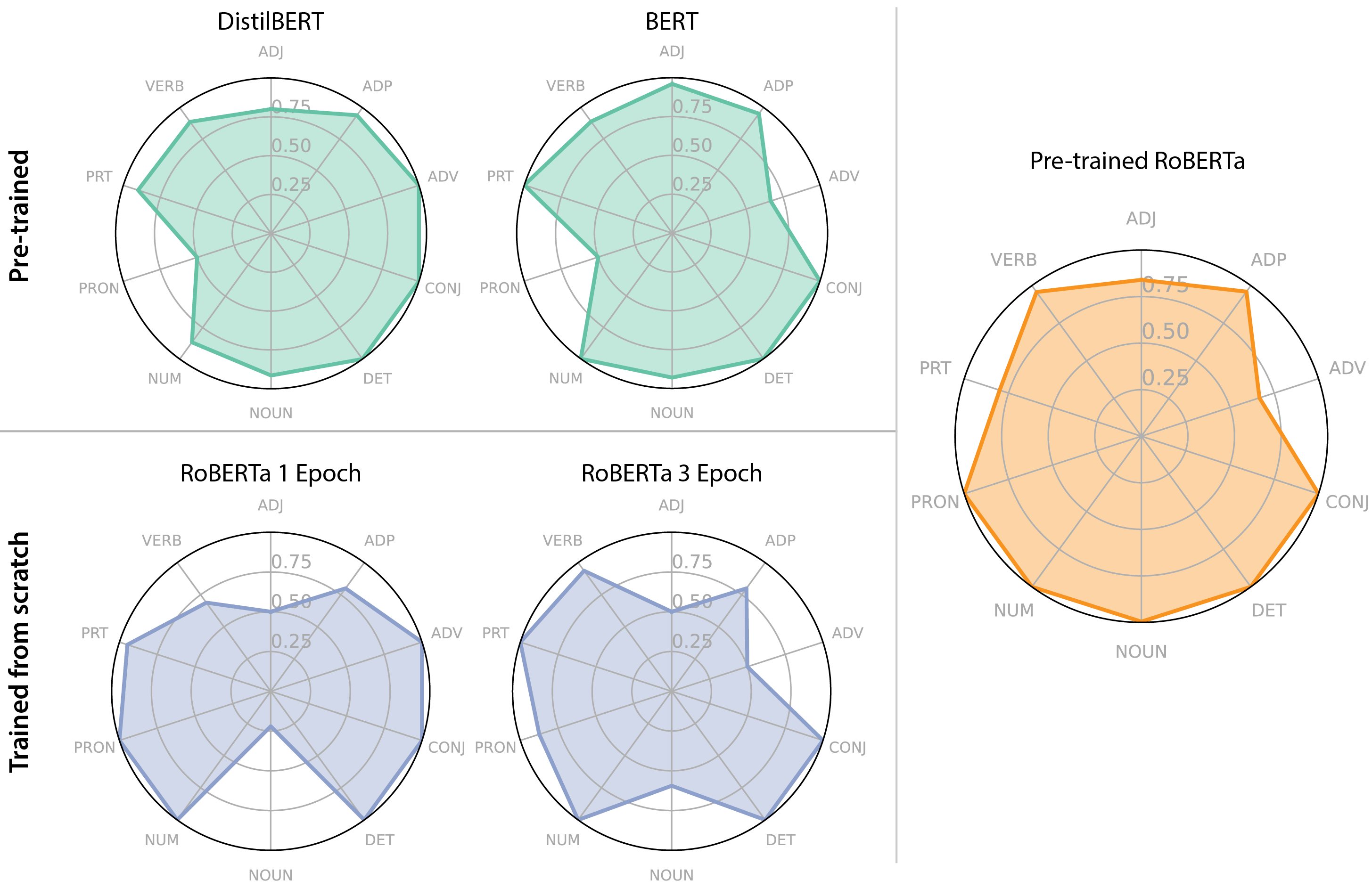}
  \caption{\label{fig:spider_squad}Micro-accuracy of POS classes for pretrained DistilBERT and BERT (green) and for the first two training phases of RoBERTa model (blue), in comparison with pretrained RoBERTa (orange) on the SQuAD dataset. }
\end{figure*}

\begin{table*}
    \centering
    \small
    \begin{tabular}{lcccccccccc} 
    \hline
    \textbf{}                                                               & \textbf{ADJ}     & \textbf{ADP}    & \textbf{ADV}     & \textbf{CONJ} & \textbf{DET} & \textbf{NOUN}    & \textbf{NUM}     & \textbf{PRON} & \textbf{PRT} & \textbf{VERB}    \\ 
    \hline
    \textbf{Ground Truth} & 5.1\%               & 21.3\%               & 0.9\% & 0.4\%             & 1.6\%            & \textbf{50.7\%}            & 3.3\%                & \textbf{0.2\%}           & 2.0\%            & 14.5\%  \\ 
    \hline
    \end{tabular}
    \caption{\label{ref:pos_frequencies} Ground truth frequencies (\% of entries in the ground truth KG) for each POS tag in the SQuAD cloze statement dataset. The smallest and largest frequencies are bolded, and the ground truth KG has 451 words in total.}
\end{table*}

\begin{table*}
    \centering
    \small
    \begin{tabular}{lcccccccccc} 
    \hline
    \textbf{Model}                                                               & \textbf{ADJ}     & \textbf{ADP}    & \textbf{ADV}     & \textbf{CONJ} & \textbf{DET} & \textbf{NOUN}    & \textbf{NUM}     & \textbf{PRON} & \textbf{PRT} & \textbf{VERB}    \\ 
    \hline
    \textbf{BERT} & -4               & 5               & \textbf{-33.3} & 0             & 0            & 6.9            & 0                & -50           & 0            & \textbf{11.2}  \\ 
    
    \textbf{DistilBERT} & \textbf{20}      & 6               & 0                & 0             & 0            & 8.5            & \textbf{-13.3} & 50            & 10           & 11.2           \\ 
    
    \textbf{RoBERTa 1e}                                                              & -49.9          & \textbf{19.9} & 0                & 0             & 0            & \textbf{-77.9} & 0                & 0             & -5.2       & -31.1          \\ 
    
    \textbf{RoBERTa 3e}                                                              & \textbf{-49.9} & 20          & \textbf{49.9}  & 0             & 0            & -40.5          & 0                & 12.5        & 0            & -6.3           \\ 
    
    \textbf{RoBERTa }   & \textbf{16}      & 4               & \textbf{-33.3} & 0             & 0            & -0.3           & 0                & 0             & -20          & 4.2            \\
    \hline
    \end{tabular}
    \caption{\label{ref:squad_pos_results} POS tagging experiment results (percent difference from ground truth for a specific POS tag category) on pretrained models (BERT, DistilBERT, and RoBERTa) and trained RoBERTa (1 epoch, 3 epochs) for SQuAD.}
\end{table*}

\subsection{Results}

In the following section, we provide experimental results addressing the two research directions of this work: quantitatively comparing language model knowledge and identifying linguistic skills over time. Accordingly, we analyze the knowledge graph extracts through the lens of quantitative graph metrics (graph-edit-distance and graph2vec similarity) and linguistics (part-of-speech tagging). Knowledge graphs were extracted across 3 pretrained models (DistilBERT, BERT, RoBERTa) and 4 trained-from-scratch models (RoBERTa epochs 1, 3, 5, and 7), through 4 cloze datasets (SQuAD, 3 Google-RE variants). We discuss interesting result highlights here; more results can be viewed in the Appendix. 

\subsubsection{How can we quantitatively compare language model knowledge acquisition?}

In Table \ref{tab:graph_results} we show graph comparison scores for 3 pretrained BERT variants and 4 trained RoBERTa models across the Google-RE Place of Birth dataset. As discussed in the previous sections, we compute graph similarity between all the models and pretrained RoBERTa using two metrics: an approximation of the graph-edit-distance (Assignment Edit Distance) between extracted graphs and the euclidean distance between the embeddings of the graphs \cite{riesen2009approximate, narayanan2017graph2vec}. Notice that BERT and DistilBERT are consistently closer to RoBERTa compared to the trained-from-scratch models (as evidenced by smaller graph-edit-distance and euclidean distance scores). Also note that the both the graph-edit-distance metric (141.25, 135, 130.50, 121.50) and the Euclidean distance of the Graph2Vec representations (0.2260, 0.1733, 0.1607, 0.1605) decrease as the epochs of trained-from-scratch RoBERTa increases. Additional manually calculated graph-edit-distance results are showcased in Appendix Tables \ref{tab:ged_score2} and \ref{tab:ged_score3}. This analysis is important because it enables us to quantitatively compare the knowledge a language model has across model variations and training stages.

\subsubsection{As a language model trains, what linguistic traits does it learn over time?}
In Table \ref{ref:squad_pos_results} and Appendix Table \ref{tab:pos_results}, we present the results of our Part of Speech (POS)-tag metrics, comparing each of the language model KGs to the ground truth KGs and identifying which parts-of-speech are being overpredicted and underpredicted. If a language model KG was performing perfectly, all of the $POSOR$ percentage differences in Table~\ref{ref:squad_pos_results} would be 0. Table \ref{ref:pos_frequencies} sets the scene for the LAMA SQuAD dataset, extracting the parts-of-speech for the ground truth knowledge graph to contextualize the results. \textit{Nouns} are most present with 50.7\% of the SQuAD data and \textit{pronouns} are least present with 0.2\% of the data. In the case of RoBERTa 1 epoch on the SQuAD dataset, we see that it is severely underpredicting nouns (-77.9\%) and overpredicting prepositions (19.9\%). In RoBERTa 3 epochs on the SQuAD dataset, we see that we are no longer as drastically underpredicting nouns (-40.6\%) and are overpredicting adverbs more than prepositions (49.9\%). 

Observing the relationships between the pretrained DistilBERT, BERT, and RoBERTa on the Google-RE datasets in Tables \ref{ref:squad_pos_results} and \ref{tab:pos_results}, we see that BERT and DistilBERT are very similar (noting that DistilBERT is a distilled form of our BERT-base-cased model). In the Google-RE Place of Birth dataset, the difference in BERT and DistilBERT's largest weaknesses (predicting numbers) and strengths (predicting particles) are less than half of a percent (-25.8\% vs. -25.7\%, 16.8\% vs. 16.3\%). In the Google-RE Place of Death dataset, we see that pretrained RoBERTa is much more accurate at predicting pronouns than BERT (0\% vs. -33.3\%). POS annotations extracted from the KG creation pipeline allow us linguistic granularity in comparing models across various epochs and variants. It also enables machine learning practitioners to produce specialized subsets of data to fix language model linguistic weaknesses in future training iterations.

Using the generated knowledge graphs, we can examine how certain relations change over time. In the RoBERTa trained for 1 epoch, a new relation states \{\textit{teachers, follow, as curricula}\}. In the 3 epoch version, that same triple changes to \{\textit{teachers, follow, curricula}\}, which demonstrates a qualitative example of a better grasp of adverbs. In the 1 epoch RoBERTa on the SQuAD dataset, 23 of 40 predicted target objects in the new LM relations were articles (e.g. in, on, the). In 3 epoch RoBERTa, we see a marked reduction in article prediction, with only 9 of 40 predicted target objects as articles.

Lastly, the radio graphs in Figure \ref{fig:spider_squad} demonstrate the POS results of RoBERTa at 1 and 3 epochs, as well as pretrained DistilBERT and BERT in comparison with pretrained RoBERTa. As the overall area of the plots increase, we note that the accuracy of the model performance in the corresponding part-of-speech categories increases as well. These analyses are conducted as a proof-of-concept on early stages of RoBERTa and pretrained BERT variants, but the metrics introduced here are broadly applicable and useful for comparing any set of language models.

\section{Conclusion}
In this paper, we present a pipeline to extract Knowledge Graphs (KGs) from masked language models using cloze statements, and 3 metrics to analyze these graphs comparatively. Our main contribution is the novelty of comparing KG extracts over time to showcase comparative insights between BERT variants at different stages of training. We are inspired by prior work from \citet{petroni2019language} to develop our extraction pipeline and expand further on their work through the lens of interpretability and a simpler relation extraction methodology. Our results demonstrate that in each training phase, language models tend to create better connections and more meaningful triples across their extracted knowledge graph, which justifies our intuition towards the performance increase across our metrics in each training epoch (RoBERTa 1 through 7 epochs) and pretrained model advancement (DistilBERT, BERT, RoBERTa). These generated knowledge graphs are a large step towards addressing the research questions: How well does my language model perform in comparison to another (using metrics other than accuracy)? What are the linguistic strengths of my language model? What kind of data should I train my model on to improve it further? Our pipeline aims to become a diagnostic benchmark for language models, providing an alternate approach for AI practitioners to identify language model strengths and weaknesses during the model training process itself.

\subsection{Future Work}
\label{ref:future_work}
This work can most certainly be extended further into the realm of other language models and other domains. If we had more resources, a natural extension would be to train RoBERTa from scratch at much larger epochs (i.e. 50, 500, 5000) to better understand later stages of the model training process, instead of looking at only the first 7 epochs. We skimmed the surface of probing task literature to identify skills present in the query answering; delving further into \citet{kim-etal-2019-probing}'s benchmarks and expanding on our POS tagging work could provide more interesting insights. Additionally, switching our model task from masking to sentence generation and feeding it into the same KG extraction pipeline could reveal alternative conclusions about grammar and linguistic strengths. Improving the dataset size and variation is another important step; more diverse data extracted from Wikipedia could provide stronger evaluation-based claims \cite{wang2021wikigraphs}.

\subsection{Acknowledgements}
This work was performed under the EPFL EDIC Fellowship at the Machine Learning and Optimization (MLO) Laboratory. We thank Matteo Pagliardini for his valuable advice on the project.

\bibliographystyle{unsrtnat}
\bibliography{custom}

\begin{thebibliography}{42}
\providecommand{\natexlab}[1]{#1}
\providecommand{\url}[1]{\texttt{#1}}
\expandafter\ifx\csname urlstyle\endcsname\relax
  \providecommand{\doi}[1]{doi: #1}\else
  \providecommand{\doi}{doi: \begingroup \urlstyle{rm}\Url}\fi

\bibitem[Devlin et~al.(2018)Devlin, Chang, Lee, and Toutanova]{devlin2018BERT}
Jacob Devlin, Ming-Wei Chang, Kenton Lee, and Kristina Toutanova.
\newblock BERT: Pre-training of deep bidirectional transformers for language
  understanding.
\newblock \emph{arXiv preprint arXiv:1810.04805}, 2018.

\bibitem[Wang et~al.(2018)Wang, Singh, Michael, Hill, Levy, and
  Bowman]{wang2018glue}
Alex Wang, Amanpreet Singh, Julian Michael, Felix Hill, Omer Levy, and Samuel~R
  Bowman.
\newblock GLUE: A multi-task benchmark and analysis platform for natural
  language understanding.
\newblock \emph{arXiv preprint arXiv:1804.07461}, 2018.

\bibitem[Rajpurkar et~al.(2016)Rajpurkar, Zhang, Lopyrev, and
  Liang]{rajpurkar2016squad}
Pranav Rajpurkar, Jian Zhang, Konstantin Lopyrev, and Percy Liang.
\newblock Squad: 100,000+ questions for machine comprehension of text.
\newblock \emph{arXiv preprint arXiv:1606.05250}, 2016.

\bibitem[Williams et~al.(2017)Williams, Nangia, and Bowman]{williams2017broad}
Adina Williams, Nikita Nangia, and Samuel~R Bowman.
\newblock A broad-coverage challenge corpus for sentence understanding through
  inference.
\newblock \emph{arXiv preprint arXiv:1704.05426}, 2017.

\bibitem[Sanh et~al.(2019)Sanh, Debut, Chaumond, and Wolf]{sanh2019distilBERT}
Victor Sanh, Lysandre Debut, Julien Chaumond, and Thomas Wolf.
\newblock DistilBERT, a distilled version of BERT: smaller, faster, cheaper and
  lighter.
\newblock \emph{arXiv preprint arXiv:1910.01108}, 2019.

\bibitem[Liu et~al.(2019)Liu, Ott, Goyal, Du, Joshi, Chen, Levy, Lewis,
  Zettlemoyer, and Stoyanov]{liu2019RoBERTa}
Yinhan Liu, Myle Ott, Naman Goyal, Jingfei Du, Mandar Joshi, Danqi Chen, Omer
  Levy, Mike Lewis, Luke Zettlemoyer, and Veselin Stoyanov.
\newblock RoBERTa: A robustly optimized BERT pretraining approach.
\newblock \emph{arXiv preprint arXiv:1907.11692}, 2019.

\bibitem[Ruder(2020)]{MLandNLP37:online}
Sebastian Ruder.
\newblock ML and NLP research highlights of
  2020.
\newblock
  \url{https://ruder.io/research-highlights-2020/index.html#5-evaluation-beyond-accuracy},
  2020.
\newblock (Accessed on 02/08/2021).

\bibitem[Fu et~al.(2020)Fu, Liu, and Neubig]{fu-etal-2020-interpretable}
Jinlan Fu, Pengfei Liu, and Graham Neubig.
\newblock Interpretable multi-dataset evaluation for named entity recognition.
\newblock In \emph{Proceedings of the 2020 Conference on Empirical Methods in
  Natural Language Processing (EMNLP)}, pages 6058--6069, Online, November
  2020. Association for Computational Linguistics.
\newblock \doi{10.18653/v1/2020.emnlp-main.489}.
\newblock URL \url{https://www.aclweb.org/anthology/2020.emnlp-main.489}.

\bibitem[Kaushik et~al.(2020)Kaushik, Hovy, and Lipton]{Kaushik2020Learning}
Divyansh Kaushik, Eduard Hovy, and Zachary Lipton.
\newblock Learning the difference that makes a difference with
  counterfactually-augmented data.
\newblock In \emph{International Conference on Learning Representations}, 2020.
\newblock URL \url{https://openreview.net/forum?id=Sklgs0NFvr}.

\bibitem[Gardner et~al.(2020)Gardner, Artzi, Basmova, Berant, Bogin, Chen,
  Dasigi, Dua, Elazar, Gottumukkala, Gupta, Hajishirzi, Ilharco, Khashabi, Lin,
  Liu, Liu, Mulcaire, Ning, Singh, Smith, Subramanian, Tsarfaty, Wallace,
  Zhang, and Zhou]{gardner2020evaluating}
Matt Gardner, Yoav Artzi, Victoria Basmova, Jonathan Berant, Ben Bogin, Sihao
  Chen, Pradeep Dasigi, Dheeru Dua, Yanai Elazar, Ananth Gottumukkala, Nitish
  Gupta, Hanna Hajishirzi, Gabriel Ilharco, Daniel Khashabi, Kevin Lin,
  Jiangming Liu, Nelson~F. Liu, Phoebe Mulcaire, Qiang Ning, Sameer Singh,
  Noah~A. Smith, Sanjay Subramanian, Reut Tsarfaty, Eric Wallace, Ally Zhang,
  and Ben Zhou.
\newblock Evaluating models' local decision boundaries via contrast sets, 2020.

\bibitem[Warstadt et~al.(2020)Warstadt, Parrish, Liu, Mohananey, Peng, Wang,
  and Bowman]{warstadts_doi}
Alex Warstadt, Alicia Parrish, Haokun Liu, Anhad Mohananey, Wei Peng, Sheng-Fu
  Wang, and Samuel~R. Bowman.
\newblock Blimp: The benchmark of linguistic minimal pairs for english.
\newblock \emph{Transactions of the Association for Computational Linguistics},
  8:\penalty0 377--392, 2020.
\newblock \doi{10.1162/tacl\_a\_00321}.
\newblock URL \url{https://doi.org/10.1162/tacl_a_00321}.

\bibitem[Petroni et~al.(2019)Petroni, Rockt{\"a}schel, Lewis, Bakhtin, Wu,
  Miller, and Riedel]{petroni2019language}
Fabio Petroni, Tim Rockt{\"a}schel, Patrick Lewis, Anton Bakhtin, Yuxiang Wu,
  Alexander~H Miller, and Sebastian Riedel.
\newblock Language models as knowledge bases?
\newblock In \emph{Proceedings of the 2019 Conference on Empirical Methods in
  Natural Language Processing (EMNLP), 2019}, 2019.

\bibitem[Wolf et~al.(2020)Wolf, Debut, Sanh, Chaumond, Delangue, Moi, Cistac,
  Rault, Louf, Funtowicz, Davison, Shleifer, von Platen, Ma, Jernite, Plu, Xu,
  Scao, Gugger, Drame, Lhoest, and Rush]{wolf-etal-2020-transformers}
Thomas Wolf, Lysandre Debut, Victor Sanh, Julien Chaumond, Clement Delangue,
  Anthony Moi, Pierric Cistac, Tim Rault, Rémi Louf, Morgan Funtowicz, Joe
  Davison, Sam Shleifer, Patrick von Platen, Clara Ma, Yacine Jernite, Julien
  Plu, Canwen Xu, Teven~Le Scao, Sylvain Gugger, Mariama Drame, Quentin Lhoest,
  and Alexander~M. Rush.
\newblock Transformers: State-of-the-art natural language processing.
\newblock In \emph{Proceedings of the 2020 Conference on Empirical Methods in
  Natural Language Processing: System Demonstrations}, pages 38--45, Online,
  October 2020. Association for Computational Linguistics.
\newblock URL \url{https://www.aclweb.org/anthology/2020.emnlp-demos.6}.

\bibitem[Rogers et~al.(2020)Rogers, Kovaleva, and Rumshisky]{rogers2020primer}
Anna Rogers, Olga Kovaleva, and Anna Rumshisky.
\newblock A primer in BERTology: What we know about how BERT works.
\newblock \emph{Transactions of the Association for Computational Linguistics},
  8:\penalty0 842--866, 2020.

\bibitem[Hewitt and Manning(2019)]{hewitt-manning-2019-structural}
John Hewitt and Christopher~D. Manning.
\newblock {A} structural probe for finding syntax in word representations.
\newblock In \emph{Proceedings of the 2019 Conference of the North {A}merican
  Chapter of the Association for Computational Linguistics: Human Language
  Technologies, Volume 1 (Long and Short Papers)}, pages 4129--4138,
  Minneapolis, Minnesota, June 2019. Association for Computational Linguistics.
\newblock \doi{10.18653/v1/N19-1419}.
\newblock URL \url{https://www.aclweb.org/anthology/N19-1419}.

\bibitem[Clark et~al.(2019)Clark, Khandelwal, Levy, and Manning]{clark2019does}
Kevin Clark, Urvashi Khandelwal, Omer Levy, and Christopher~D. Manning.
\newblock What does BERT look at? an analysis of BERT's attention, 2019.

\bibitem[Goldberg(2019)]{goldberg2019assessing}
Yoav Goldberg.
\newblock Assessing BERT's syntactic abilities, 2019.

\bibitem[Coenen et~al.(2019)Coenen, Reif, Yuan, Kim, Pearce, Vi{\'e}gas, and
  Wattenberg]{coenen2019visualizing}
Andy Coenen, Emily Reif, Ann Yuan, Been Kim, Adam Pearce, Fernanda Vi{\'e}gas,
  and Martin Wattenberg.
\newblock Visualizing and measuring the geometry of BERT.
\newblock \emph{arXiv preprint arXiv:1906.02715}, 2019.

\bibitem[Tenney et~al.(2019)Tenney, Das, and Pavlick]{tenney2019BERT}
Ian Tenney, Dipanjan Das, and Ellie Pavlick.
\newblock BERT rediscovers the classical nlp pipeline, 2019.

\bibitem[Peters et~al.(2018)Peters, Neumann, Iyyer, Gardner, Clark, Lee, and
  Zettlemoyer]{peters-etal-2018-deep}
Matthew Peters, Mark Neumann, Mohit Iyyer, Matt Gardner, Christopher Clark,
  Kenton Lee, and Luke Zettlemoyer.
\newblock Deep contextualized word representations.
\newblock In \emph{Proceedings of the 2018 Conference of the North {A}merican
  Chapter of the Association for Computational Linguistics: Human Language
  Technologies, Volume 1 (Long Papers)}, pages 2227--2237, New Orleans,
  Louisiana, June 2018. Association for Computational Linguistics.
\newblock \doi{10.18653/v1/N18-1202}.
\newblock URL \url{https://www.aclweb.org/anthology/N18-1202}.

\bibitem[Hao et~al.(2019)Hao, Dong, Wei, and Xu]{hao2019visualizing}
Yaru Hao, Li~Dong, Furu Wei, and Ke~Xu.
\newblock Visualizing and understanding the effectiveness of BERT, 2019.

\bibitem[Liu et~al.(2021)Liu, Wang, Kasai, Hajishirzi, and
  Smith]{liu2021probing}
Leo~Z. Liu, Yizhong Wang, Jungo Kasai, Hannaneh Hajishirzi, and Noah~A. Smith.
\newblock Probing across time: What does roBERTa know and when?, 2021.

\bibitem[Chiang et~al.(2020)Chiang, Huang, and yi~Lee]{chiang2020pretrained}
Cheng-Han Chiang, Sung-Feng Huang, and Hung yi~Lee.
\newblock Pretrained language model embryology: The birth of alBERT, 2020.

\bibitem[Ribeiro et~al.(2020)Ribeiro, Wu, Guestrin, and
  Singh]{ribeiro-etal-2020-beyond}
Marco~Tulio Ribeiro, Tongshuang Wu, Carlos Guestrin, and Sameer Singh.
\newblock Beyond accuracy: Behavioral testing of {NLP} models with
  {C}heck{L}ist.
\newblock In \emph{Proceedings of the 58th Annual Meeting of the Association
  for Computational Linguistics}, pages 4902--4912, Online, July 2020.
  Association for Computational Linguistics.
\newblock \doi{10.18653/v1/2020.acl-main.442}.
\newblock URL \url{https://www.aclweb.org/anthology/2020.acl-main.442}.

\bibitem[Petroni et~al.(2020)Petroni, Lewis, Piktus, Rockt{\"a}schel, Wu,
  Miller, and Riedel]{petroni2020how}
Fabio Petroni, Patrick Lewis, Aleksandra Piktus, Tim Rockt{\"a}schel, Yuxiang
  Wu, Alexander~H. Miller, and Sebastian Riedel.
\newblock How context affects language models' factual predictions.
\newblock In \emph{Automated Knowledge Base Construction}, 2020.
\newblock URL \url{https://openreview.net/forum?id=025X0zPfn}.

\bibitem[Heinzerling and Inui(2021)]{heinzerling2021language}
Benjamin Heinzerling and Kentaro Inui.
\newblock Language models as knowledge bases: On entity representations,
  storage capacity, and paraphrased queries, 2021.

\bibitem[Jiang et~al.(2020)Jiang, Xu, Araki, and Neubig]{jiang2020can}
Zhengbao Jiang, Frank~F Xu, Jun Araki, and Graham Neubig.
\newblock How can we know what language models know?
\newblock \emph{Transactions of the Association for Computational Linguistics},
  8:\penalty0 423--438, 2020.

\bibitem[Aspillaga et~al.(2021)Aspillaga, Mendoza, and
  Soto]{aspillaga2021tracking}
Carlos Aspillaga, Marcelo Mendoza, and Alvaro Soto.
\newblock Tracking the progress of language models by extracting their
  underlying knowledge graphs, 2021.
\newblock URL \url{https://openreview.net/forum?id=ghKbryXRRAB}.

\bibitem[Goswami et~al.(2020)Goswami, Bhat, Ohana, and
  Rekatsinas]{goswami2020unsupervised}
Ankur Goswami, Akshata Bhat, Hadar Ohana, and Theodoros Rekatsinas.
\newblock Unsupervised relation extraction from language models using
  constrained cloze completion, 2020.

\bibitem[Sanfeliu and Fu(1983)]{grapheditdistance}
AlBERTo Sanfeliu and King-Sun Fu.
\newblock A distance measure between attributed relational graphs for pattern
  recognition.
\newblock \emph{IEEE Transactions on Systems, Man, and Cybernetics},
  SMC-13\penalty0 (3):\penalty0 353--362, 1983.
\newblock \doi{10.1109/TSMC.1983.6313167}.

\bibitem[Narayanan et~al.(2017)Narayanan, Chandramohan, Venkatesan, Chen, Liu,
  and Jaiswal]{narayanan2017graph2vec}
Annamalai Narayanan, Mahinthan Chandramohan, Rajasekar Venkatesan, Lihui Chen,
  Yang Liu, and Shantanu Jaiswal.
\newblock graph2vec: Learning distributed representations of graphs.
\newblock \emph{arXiv preprint arXiv:1707.05005}, 2017.

\bibitem[Kim et~al.(2019)Kim, Patel, Poliak, Xia, Wang, McCoy, Tenney, Ross,
  Linzen, Van~Durme, Bowman, and Pavlick]{kim-etal-2019-probing}
Najoung Kim, Roma Patel, Adam Poliak, Patrick Xia, Alex Wang, Tom McCoy, Ian
  Tenney, Alexis Ross, Tal Linzen, Benjamin Van~Durme, Samuel~R. Bowman, and
  Ellie Pavlick.
\newblock Probing what different {NLP} tasks teach machines about function word
  comprehension.
\newblock In \emph{Proceedings of the Eighth Joint Conference on Lexical and
  Computational Semantics (*{SEM} 2019)}, pages 235--249, Minneapolis,
  Minnesota, June 2019. Association for Computational Linguistics.
\newblock URL \url{https://www.aclweb.org/anthology/S19-1026}.

\bibitem[Honnibal et~al.(2020)Honnibal, Montani, Van~Landeghem, and
  Boyd]{spacy}
Matthew Honnibal, Ines Montani, Sofie Van~Landeghem, and Adriane Boyd.
\newblock {spaCy: Industrial-strength Natural Language Processing in Python},
  2020.
\newblock URL \url{https://doi.org/10.5281/zenodo.1212303}.

\bibitem[tex()]{textacy47:online}
Textacy library.
\newblock \url{https://pypi.org/project/textacy/}.
\newblock (Accessed on 02/10/2021).

\bibitem[Goo()]{GoogleAI29:online}
50,000 lessons on how to read: a relation extraction corpus.
\newblock
  \url{https://ai.googleblog.com/2013/04/50000-lessons-on-how-to-read-relation.html}.
\newblock (Accessed on 02/08/2021).

\bibitem[Riesen and Bunke(2009)]{riesen2009approximate}
Kaspar Riesen and Horst Bunke.
\newblock Approximate graph edit distance computation by means of bipartite
  graph matching.
\newblock \emph{Image and Vision computing}, 27\penalty0 (7):\penalty0
  950--959, 2009.

\bibitem[Rozemberczki and Sarkar(2020)]{rozemberczki2020characteristic}
Benedek Rozemberczki and Rik Sarkar.
\newblock Characteristic functions on graphs: Birds of a feather, from
  statistical descriptors to parametric models.
\newblock In \emph{Proceedings of the 29th ACM International Conference on
  Information \& Knowledge Management}, pages 1325--1334, 2020.

\bibitem[Saphra and Lopez(2018)]{saphra2018language}
Naomi Saphra and Adam Lopez.
\newblock Language models learn POS first.
\newblock In \emph{Proceedings of the 2018 EMNLP Workshop BlackboxNLP:
  Analyzing and Interpreting Neural Networks for NLP}, pages 328--330, 2018.

\bibitem[Manning(2011)]{manning2011part}
Christopher~D Manning.
\newblock Part-of-speech tagging from 97\% to 100\%: is it time for some
  linguistics?
\newblock In \emph{International conference on intelligent text processing and
  computational linguistics}, pages 171--189. Springer, 2011.

\bibitem[Voutilainen(2003)]{voutilainen2003part}
Atro Voutilainen.
\newblock Part-of-speech tagging.
\newblock \emph{The Oxford handbook of computational linguistics}, pages
  219--232, 2003.

\bibitem[Loper and Bird(2002)]{nltk}
Edward Loper and Steven Bird.
\newblock Nltk: The natural language toolkit.
\newblock In \emph{Proceedings of the ACL-02 Workshop on Effective Tools and
  Methodologies for Teaching Natural Language Processing and Computational
  Linguistics - Volume 1}, ETMTNLP '02, page 63–70, USA, 2002. Association
  for Computational Linguistics.
\newblock \doi{10.3115/1118108.1118117}.
\newblock URL \url{https://doi.org/10.3115/1118108.1118117}.

\bibitem[Wang et~al.(2021)Wang, Li, Aslan, and Vinyals]{wang2021wikigraphs}
Luyu Wang, Yujia Li, Ozlem Aslan, and Oriol Vinyals.
\newblock Wikigraphs: A Wikipedia text - knowledge graph paired dataset, 2021.

\end{thebibliography}

\begin{figure*}[]
  \section*{Appendix}
  \centering
  \includegraphics[width=0.95\textwidth]{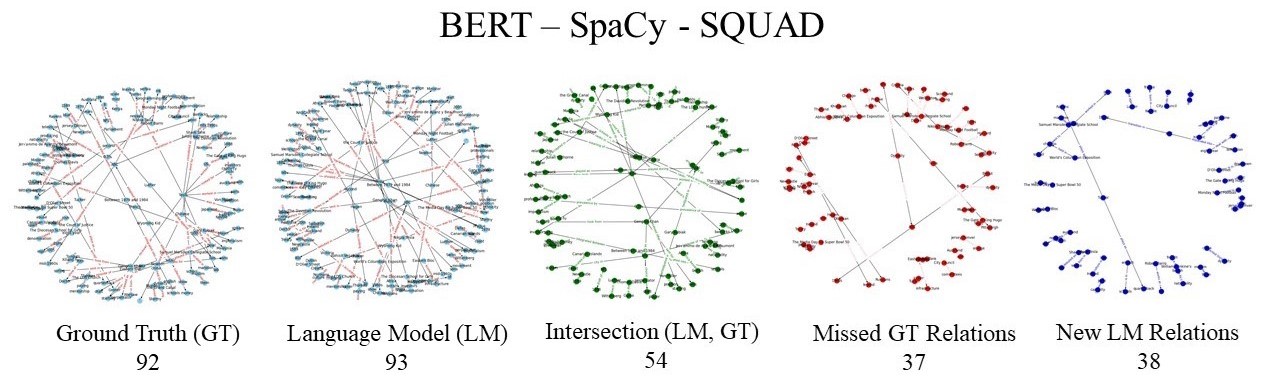}
  \includegraphics[width=0.95\textwidth]{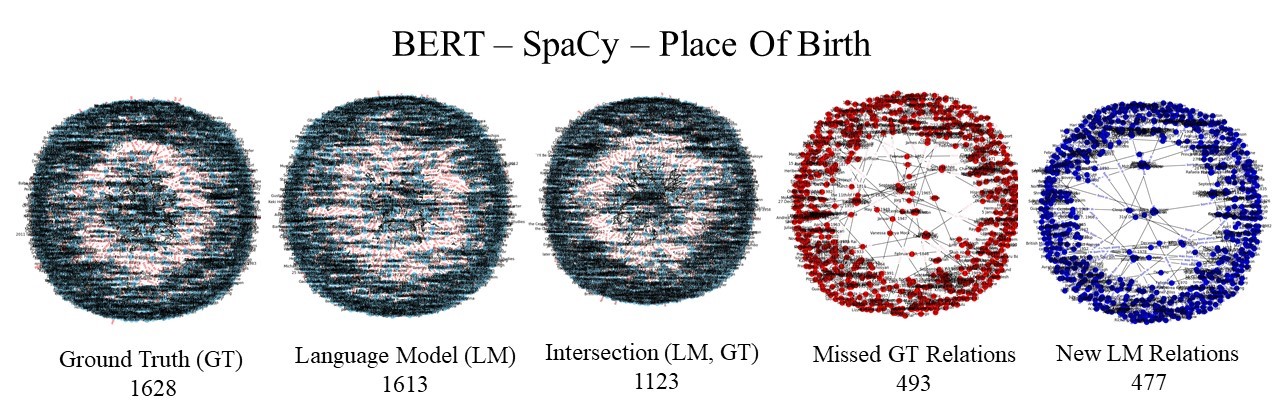}
  \includegraphics[width=0.95\textwidth]{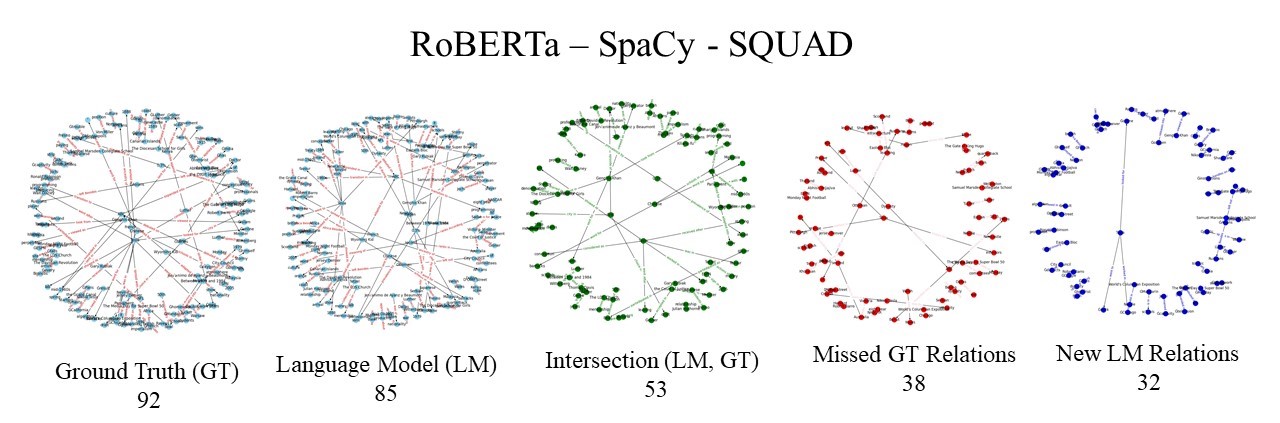}
  \includegraphics[width=0.95\textwidth]{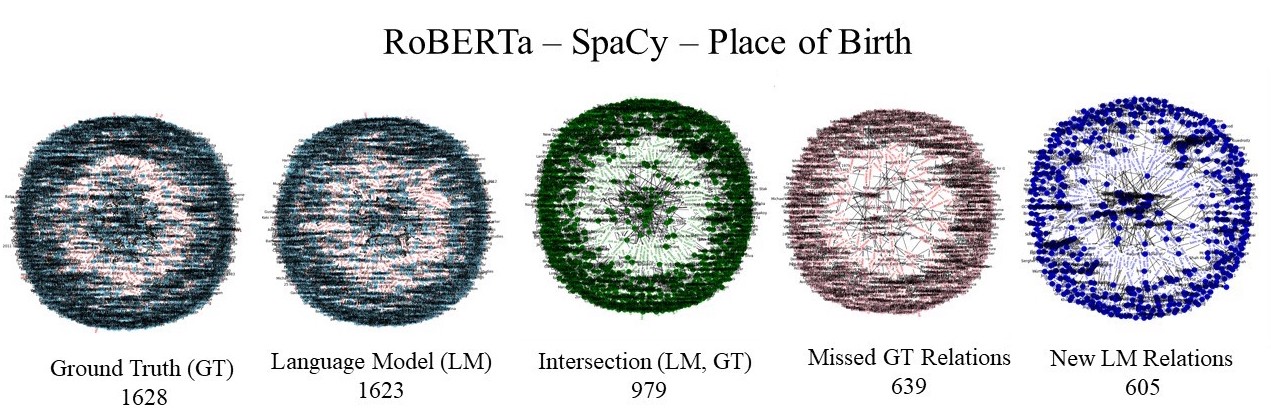}
  \caption{\label{fig:bert_squad}Knowledge Graphs (KGs) generated using the pretrained BERT / RoBERTa architecture and SpaCy relation extraction on the LAMA SQUAD and Google RE datasets. The numbers below each KG represent the number of unique relation triples. The LM KG represents the cloze sentences after the masks have been replaced by the Language Model predictions. Ground Truth references gold label sentences where relations have been extracted directly.}
\end{figure*}
\pagebreak
\begin{table*}[]
    \centering
    \begin{tabular}{lcccc}
    \hline
    \textbf{SQuAD} & \textbf{MobileBERT} & \textbf{DistilBERT} & \textbf{BERT} & \textbf{RoBERTa} \\ \hline
    \textbf{MobileBERT}                       & 0     & 15    & 21     & 25       \\
    \textbf{DistilBERT}                       & 15      & 0     & 12     & 23       \\ 
    \textbf{BERT}                          & 21   & 12     & 0     & 17      \\ 
    \textbf{RoBERTa}                      & 25       & 23     & 17     & 0      \\ \hline
    \end{tabular}
    \caption{\label{tab:ged_score2}Graph-Edit-Distance scores between KGs for BERT variants using the SQUAD dataset from LAMA \cite{petroni2019language}.}
\end{table*}

\begin{table*}[]
    \centering
    \begin{tabular}{lccc}
    \hline
    \textbf{Google-RE} & \textbf{DistilBERT} & \textbf{BERT} & \textbf{RoBERTa} \\ \hline
    \textbf{DistilBERT}                             & 0     & 113     & 272       \\ 
    \textbf{BERT}                             & 113     & 0     & 287      \\ 
    \textbf{RoBERTa}                             & 272     & 287     & 0      \\ \hline
    \end{tabular}
    \caption{\label{tab:ged_score3}Graph-Edit-Distance scores between KGs for BERT variants using the Google-RE Place of Birth Dataset from LAMA \cite{petroni2019language}.}
\end{table*}

\begin{table*}
\centering
\small
\begin{tabular}{lcccccc} 
\hline
POS-tag Experiment                         & \textbf{.}      & \textbf{ADJ}     & \textbf{ADP}     & \textbf{ADV}     & \textbf{CONJ} & \textbf{DET}     \\ 
\hline
\textbf{re-date-birth BERT}                & \textbf{-5.3} & -1.1           & 0.6            & 0                & 0             & 6.2             \\ 

\textbf{re-date-birth RoBERTa 1e}          & 0               & -6.2            & 0                & 0                & 0             & \textbf{66.6}  \\ 

\textbf{re-date-birth RoBERTa 3e}          & 0               & -6.2            & \textbf{-33.3} & 0                & 0             & \textbf{53.8}  \\ 

\textbf{re-date-birth RoBERTa (pretrain)}  & \textbf{52.6} & -1.7           & 0.9            & 5                & 9         & 2            \\ 

\textbf{re-place-birth BERT}               & 0.5           & -0.4           & -2.1           & 3.1            & 0             & 2.5              \\ 

\textbf{re-place-birth DistilBERT}         & 1.5           & -1.7           & -2.1           & 3.1            & 0             & 1.6            \\ 

\textbf{re-place-birth RoBERTa 1e}         & 0               & -4               & \textbf{-37.4} & 0                & 0             & 0                \\ 

\textbf{re-place-birth RoBERTa 3e}         & 0               & -4               & \textbf{-37.4} & 0                & 0             & 0                \\ 

\textbf{re-place-birth RoBERTa (pretrain)} & -2.5          & -8.9           & 3.6            & -3.1           & 3.4         & 1.6            \\ 

\textbf{re-place-death BERT}               & 20.6          & 19.2           & \textbf{27.1}  & -19          & 14.2        & 20.5           \\ 

\textbf{re-place-death DistilBERT}         & 22.2          & 19.2           & \textbf{27.9}  & \textbf{-14.2} & 14.2        & 20.5           \\ 

\textbf{re-place-death RoBERTa (pretrain)} & 20.6          & 15.3           & \textbf{33}  & \textbf{-19} & 14.2        & 17.6           \\ 

\textbf{squad BERT}                        & 0               & -4               & 5                & \textbf{-33.3} & 0             & 0                \\ 

\textbf{squad DistilBERT}                  & 0               & \textbf{20}      & 6                & 0                & 0             & 0                \\ 

\textbf{squad RoBERTa 1e}                  & 0               & -49.9          & \textbf{19.9}  & 0                & 0             & 0                \\ 

\textbf{squad RoBERTa 3e}                  & 0               & \textbf{-49.9} & 19.9           & \textbf{49.9}  & 0             & 0                \\ 

\textbf{squad RoBERTa (pretrain)}          & 0               & \textbf{16}      & 4                & \textbf{-33.3} & 0             & 0                \\
\hline
\end{tabular}
\end{table*}

\begin{table*}
\centering
\small
\begin{tabular}{lccccc} 
\hline

POS-tag Experiment                         & \textbf{NOUN}    & \textbf{NUM}     & \textbf{PRON}    & \textbf{PRT}    & \textbf{VERB}    \\ 
\hline
\textbf{re-date-birth BERT}                & -1.9           & \textbf{14.4}  & 0                & -1.8          & 1.9            \\ 
\textbf{re-date-birth DistilBERT}          & -0.6           & 8            & 0                & 1.8           & 1.2            \\ 

\textbf{re-date-birth RoBERTa 1e}          & 12.4           & \textbf{-74.9} & -0.8           & 0               & -3.6           \\ 
\textbf{re-date-birth RoBERTa 3e}          & 5.9            & 0                & 0                & 0               & -9.2           \\ 
\textbf{re-date-birth RoBERTa (pretrain)}  & \textbf{-7.6}  & 60               & 0                & -1.2           & 9.2            \\ 
\textbf{re-place-birth BERT}               & -2.2           & \textbf{-25.7} & 14.2           & \textbf{16.8} & -1.7            \\ 
\textbf{re-place-birth DistilBERT}         & -1.9           & \textbf{-25.7} & 14.2           & \textbf{16.3} & -1.9           \\ 
\textbf{re-place-birth RoBERTa 1e}         & -0.5           & 0                & 0                & \textbf{5.5}  & 0.7            \\ 
\textbf{re-place-birth RoBERTa 3e}         & -1.2           & 0                & 0                & \textbf{5.5}  & -0.3           \\ 
\textbf{re-place-birth RoBERTa (pretrain)} & 3.9            & \textbf{-24.3} & 14.2           & \textbf{17.3} & 3.5            \\ 
\textbf{re-place-death BERT}               & 24.2           & 12.2           & \textbf{-33.3} & 17.4           & 26.5           \\ 
\textbf{re-place-death DistilBERT}         & 25.2           & 13.6           & 0                & 17.4           & 26.8           \\ 
\textbf{re-place-death RoBERTa (pretrain)} & 28.8           & 19.3           & 0                & 19          & 31.5           \\ 
\textbf{squad BERT}                        & 6.9            & 0                & -50              & 0               & \textbf{11.2}  \\ 
\textbf{squad DistilBERT}                  & 8.5            & \textbf{-13.3} & 50               & 10              & 11.2           \\ 
\textbf{squad RoBERTa 1e}                  & \textbf{-77.9} & 0                & 0                & -5.2          & -31.1          \\ 
\textbf{squad RoBERTa 3e}                  & -40.5          & 0                & 12.4           & 0               & -6.3           \\ 
\textbf{squad RoBERTa (pretrain)}          & -0.3           & 0                & 0                & -20             & 4.2            \\
\hline
\end{tabular}
\vspace{3mm}
\caption{\label{tab:pos_results}Difference in POS tags between the language-model-generated KG and ground truth KG. The largest positive and negative difference are highlighted. POS-tags use the NLTK Averaged Perceptron Tagger abbreviations: VERB - verbs (all tenses and modes), NOUN - nouns (common and proper), PRON - pronouns, ADJ - adjectives, ADV - adverbs, ADP - adpositions (prepositions and postpositions), CONJ - conjunctions, DET - determiners, NUM - cardinal numbers, PRT - particles or other function words, . - punctuation \cite{nltk}.}
\end{table*}

\end{document}